# Industrial Engineering & Management

**Research Article** **Open Access**

# Factor Analysis in Fault Diagnostics Using Random Forest

**Nagdev Amruthnath\* and Tarun Gupta**

*Industrial and Entrepreneurial Engineering, Western Michigan University, Kalamazoo, MI, 49008, USA*

## Abstract

Factor analysis or sometimes referred to as variable analysis has been extensively used in classification problems for identifying specific factors that are significant to particular classes. This type of analysis has been widely used in application such as customer segmentation, medical research, network traffic, image, and video classification. Today, factor analysis is prominently being used in fault diagnosis of machines to identify the significant factors and to study the root cause of a specific machine fault. The advantage of performing factor analysis in machine maintenance is to perform prescriptive analysis (helps answer what actions to take?) and preemptive analysis (helps answer how to eliminate the failure mode?). In this paper, a real case of an industrial rotating machine was considered where vibration and ambient temperature data was collected for monitoring the health of the machine. Gaussian mixture model-based clustering was used to cluster the data into significant groups, and spectrum analysis was used to diagnose each cluster to a specific state of the machine. The significant features that attribute to a particular mode of the machine were identified by using the random forest classification model. The significant features for specific modes of the machine were used to conclude that the clusters generated are distinct and have a unique set of significant features.

## Introduction

Fault diagnosis has been one of the critical components in predictive maintenance. In manufacturing, when the machine fails, most of the maintenance time is spent towards investigating the failure through the process of trial and error or experience. Even in cases where the fault is detected early, using detection techniques; in many cases, the time to investigate the failure is much higher than actual maintenance time. With early failure detection, the maintenance could be performed during the non-production time which may reduce the cost of lost production time but, it still increases the maintenance cost. In one benchmark survey from approximately 50 major corporations worldwide in the early 1990s, it was found that preventive maintenance budget/cost attributed to 15% to 18% of the total cost and predictive maintenance budget/cost associated to 10% to 12% [1]. Some of the other Industry O&M metrics and benchmarks are as shown in Table 1. Most organizations usually have a common goal when it comes to maintenance such as increasing the availability and reliability of the machines while decreasing the maintenance cost [2]. In the recent years, high maintenance costs have been driving manufacturing industries from total preventive maintenance (TPM) to predictive maintenance (also called condition-based monitoring) where the maintenance on a machine is performed when it is needed. Another disadvantage of preventive maintenance is identifying the best interval between maintenance and a technique to overcome this has been studied in the literature [3]. Not knowing the best interval can lead to failures or replacing the part too soon and hence increasing the maintenance cost.

There are various advantages to predictive maintenance that have been cited in the past such as driving the maintenance cost down, utilizing the complete life of part and increase in the production time [4]. Hence, a significant transition has been observed from "just-in-case" to "just-in-time" maintenance where critical machines are monitored continuously to observe their health and any deviation from their normal condition is the early stages of degradation. Some of the other advantages to predictive maintenance such as an increase in productivity and quality, and a decrease in product cost [5]. Lost productivity has been one of the undesirable forces in failing to achieve JIT in manufacturing which involves producing the highest quality products at the least cost in the lowest possible lead time [6]. In the age of connected devices, the Internet of Things (IoT) sensors have played a vital role in monitoring critical assets in real time across various manufacturing environments. This continuous stream of data in real-time has enabled in the creation of new technologies that are capable of performing real-time anomaly detection on time series data. The cost, reliability, and security of these sensors might also be attributed as one of the driving forces for condition-based monitoring [7].

Physics-based data, process data or a hybrid of both is one of the most sought data in condition-based monitoring. Today, different techniques such as machine learning, A.I, data mining, and statistics are used to develop software models to monitor the health of the machine continuously. There are four main components to condition-based monitoring (CBM) and are as follows:

a. Fault detection.

b. Fault classification.

**\*Corresponding author:** Nagdev Amruthnath, Industrial and Entrepreneurial Engineering, Western Michigan University, Kalamazoo, MI, 49008, USA, Tel: (269)387-1000; E-mail: nagdev.amruthnath@wmich.edu

| Metric | Benchmark |
|---|---|
| Equipment availability | >95% |
| Schedule compliance | >90% |
| Emergency maintenance percentage | <10% |
| Maintenance overtime percentage | <5% |
| Preventive maintenance completion percentage | >90% |
| Preventive maintenance budge/cost | 15-18% |
| Predictive maintenance budge/cost | 10-12% |

**Table 1:** Operation and maintenance metrics for U.S. Industries [1].

   c.   Time to failure prediction.

   d.   Factor analysis.

Fault detection is the first step to predictive maintenance. During the detection phase, machine sensor data is analyzed to determine if there is a change in the health of the machine. Some of the commonly used models in condition-based monitoring for fault detection are binary classification, PCA-$T^2$ and SPE statistic [8,9], k-means clustering [8], one class support vector machine (SVM) [10], artificial neural networks and logistic regression [11]. Extensive research has been performed in early fault detection using unsupervised learning where anomalies in machine health can be accurately detected without the need for historical health labels [8].

Time to failure prediction is an estimation method to predict the time at which the completely fails or breaks down. The theory behind estimating the time to failure relies on identifying the distribution of failure such as exponential or Weibull. There are various methods to estimate the distribution of the failure such as using plotting data, statistical tests [12], and goodness of fit. Some of the popular models used for predicting time to failures are SVM [13], non-linear models, linear regression and neural networks [14]. In recent times, forecasting techniques such as ARMA has gained much prominence in predictive maintenance.

Fault diagnosis is a process of diagnosing the failure mode of the machine [15]. For example, in a motor, the failure mode can be an imbalance, shaft displacement or bearing issues. One of the popular and most commonly used diagnoses in the past is vibration spectrum analysis [16]. The vibration data in the time domain is converted to the frequency spectrum, and this spectrum is diagnosed to identify the failure mode of the machine. Today, this is still one of the most popular techniques and widely used in manufacturing. Spectrum analysis has also evolved from vibration to sound, pressure, light, force, energy and other signals. Although, this is still one of the most popular techniques, one of the main challenges of this method is the need for high domain knowledge and experience. In most cases with hundreds of machines in a manufacturing facility, this method nearly becomes impractical and expensive. Due to this problem supervised machine learning techniques are frequently used to diagnose the faults in the machines. Some of the most commonly used classification models for fault diagnosis are multi-class SVM [17], K-nearest neighbor [18], neural networks [19,20], and decision trees [21]. In changing environment such as manufacturing, if the classification models are not trained for all states of the machine then, a new state of the machine (not part of the trained model) will be misclassified to a known state. Hence, unsupervised learning techniques have become more popular in fault state detection using clustering. Some of the commonly used techniques in clustering are the Gaussian finite mixture model [15], self-organizing map, hierarchical clustering [8] and density-based clustering.

Factor analysis is a technique that involves identifying significant factors for a particular group (or cluster). Factor analysis has been widely used in classification problems such as customer segmentation [22], cancer studies [23], clinical studies [24], and environmental studies [25]. This concept has been widely used in other problems such as regression and dimensionality reduction. Similar to customer segmentation, in maintenance, it is essential to identify the key features that attribute to a specific fault or failure mode of the machine. These specific features are used to study the root cause of a particular problem in the machine, and necessary design changes could be made to eliminate the problem. In other instances, when a fault is detected, these specific features can be used to verify the state of the machine.

Today, some of the most commonly used factor analysis techniques are ANOVA analysis [26], principal component analysis (PCA) [26], tree-based models, LASSO regression, and linear discriminant analysis. Although models mentioned above are mostly used for feature selection before model building, some of the models such as linear regression [27], xgboost and random forest models have the capability of providing variable importance after the models are trained. Recently a new technique called Local Interpretable Model-Agnostic Explanations (LIME) was presented with an objective of answering questions such as "*Why the model should be trusted ?*", "*what factors are contributing to the accuracy?*", "*what variables are important?*" [28].

In this paper, a vibration monitor was used on rotating machinery to observe the health of the machine [1]. Gaussian mixture model-based clustering was used to cluster the observations into specific groups [15]. Each group was diagnosed with failure mode using spectrum analysis. Finally, a random forest model was used to identify significant variables that affect each group or a cluster.

The two main objectives of this research were as follows:

***Objective 1:*** to develop a technique that is capable of diagnosing different states of the machine without the need for labels

***Objective 2:*** to identify the most critical factors that affect the causality of each state of the machine

To achieve these objectives, this paper provides an introduction to Gaussian mixture model and the need for it in fault diagnosis and vibration analysis to diagnose each state once. Then, the clustered data is sampled to build a random forest model and using the variable importance technique; each important factor was diagnosed for each state of the machine. In the end, sufficient rationale was provided for the need for this proposed technique in predictive maintenance architecture along with the future scope of this work.

## Gaussian Mixture Model

Clustering is a process of modeling similar observations into specific groups. This process in machine learning is an unsupervised learning technique where only predictor variables are known for grouping. There are various methods to group the data such as distance [29], density, shape, and type of data. In this research, the clustering technique was developed for performing fault diagnosis and factor analysis. The steps involved in this technique were as follows:

1. Identify the machine where the states of the machines need to be diagnosed.

2. Install appropriate sensors and collect the data continuously.

3. Extract statistics based featured and domain-specific features.

4. Identify the optimal number of states in the machine using within the sum of squares technique.

5. Cluster the data into an optimal number of clusters using GMM and EM technique.

6. Create random samples for each cluster from the overall population.

7. Split the sample data in training and test sets.

8. Build a random forest model using a train set along with different tuning parameters such as optimizing the number of trees and randomly selecting predictor variables.

9. Test the reliability of the model with test set data and identify the kappa value. If the value is reasonable, then use it to develop important variables for each machine state.

10. Use these important variables for each state to perform prescriptive and preemptive analysis.

A Gaussian Mixture Model is a parametric probability density function represented as a weighted sum of Gaussian component densities [30]. In clustering using finite mixture modeling, each component probability refers to a cluster, and the models that differ in the number of components/component distributions can be equated using statistical tests [31].

## Gaussian mixture model

In this section, the density functions, log-likelihood function and E and M steps for Gaussian Mixture Modeling are discussed [32]. Here, a D-dimensional continuous random vector $X$ $R^d$ is considered. From the reference (eqns 1-7), the probability density function for a mixture model which is a linear combination of $M$ Gaussian component densities is defined as [32]

$$p(x) = \sum_{j=1}^{M} P(j) \, p(x \mid j) \tag{1}$$

Where $P(j)$ is the mixture proportion and is non-negative. Its sum must be equal to one. The Gaussian centers can be defined by their centers $c_j$ and their covariance matrix $\sum_j$ [32].

$$p(x|j) = (2\pi)^{-\frac{d}{2}} \left| \sum_j \right|^{\frac{1}{2}} \cdot \exp\left[ -\frac{1}{2} (x - c_j)^T \sum_j^{-1} (x - c_j) \right] \tag{2}$$

Based on our above mixture model (eqn 4), we can define the log-likelihood function as (eqn. (3)) [32]

$$L(\theta) = \log \prod_{n-1}^{N} p(x_n) \tag{3}$$

Where, $\theta$ is the model parameter of $P(j)$, $c_j$ and $\sum_j$. Using Expectation and Maximization approach the maximum likelihood estimate of $\theta$ can be obtained iteratively. Expectation or E-step involves computing the expected value of some unobserved data using current parameter estimates and observed data. Maximization or M-step involves using the expected values from E-step to compute the maximum likelihood estimates. Upon achieving this model parameters are updated. At a given iteration step $t$,

E-step: [32]

$$p^{(t)}(j|x_n) = \frac{p^{(t)}(x_n \mid j) P^{(t)}(j)}{p^{(t)}(x_n)} \tag{4}$$

E-step: [32]

$$c_j^{(t+1)} = \frac{\sum_{n=1}^{N} p^{(t)}(j \mid x_n) x_n}{\sum_{n=1}^{N} P^{(t)}(j \mid x_n)} \tag{5}$$

$$\sum_j^{(t+1)} = \frac{\sum_{n=1}^{N} P^{(t)}(x_n - c_j^{(t+1)})(x_n - c_j^{(t+1)})^T}{\sum_{n=1}^{N} P^{(t)}(j|x_n)} \tag{6}$$

$$P^{(t+1)}(j) = \frac{1}{N} \sum_{n=1}^{N} P^{(t)}(j \mid x_n) \tag{7}$$

## Data collection

In this research, rotating machinery was considered which operated in a high-temperature environment. Vibration sensors were mounted on X-axis (Axial) and Y-axis (Radial) [1]. The vibration data was collected in time series at a sample rate of 2048 Hz. This data is collected every 5 minutes for five months continuously. Time signals were converted to frequency signals, and the features were extracted in both domains. Since the operating frequency for machinery was known to be 26.1 Hz, the features in the frequency domain were collected around this band. Every instance of raw data collected consists of approximately 1600 data point in the time domain. It was vital to capture different features both in the time domain and frequency domain. Here, the time domain features such as min, max, median, mean, standard deviation, kurtosis, skewness, range [33], and RMS [34] for both x-axis and y-axis were collected. The raw data in the time domain was later transformed to the frequency domain using Fourier transforms [35] to capture the same features in x-axis and y-axis at 25 Hz to minimize the noise at a higher frequency. Total of 36 vibration features and ambient temperature around this machinery were obtained.

## Cluster analysis

In most clustering models, the number of clusters to be formed is user-defined. The most commonly used techniques for finding the optimal number of clusters are by using (AIC) [36], BIC [36], within the sum of square (WSS), gap statistic or silhouette with the method. As the size of the data increases, AIC and BIC methods fail to provide the optimal number of clusters. In such cases, WSS and silhouette width are calculated for k clusters. Using the elbow method for WSS, an optimal number of clusters can be identified. For silhouette width, the kth cluster that provides the maximum separation is considered as the optimal number of clusters. In this research, both WSS and silhouette techniques are performed to identify the optimal number of clusters as shown in Figures 1-3.

From the WSS technique, we can observe that after the sixth cluster, there is no significant change in WSS. Hence, in WSS method the optimal number of clusters could be identified as six. In the silhouette technique, at the sixth cluster provided the maximum separation compared to other clusters [37]. Hence the optimal number of clusters was determined to be six. From both the techniques, we can identify that the optimal number of clusters is six. The data is clustered using GMM-EM with k=6 as an optimal number of clusters. The clustering analysis was performed using the mlcust package in R [31]. The results are as shown in Figure 2.

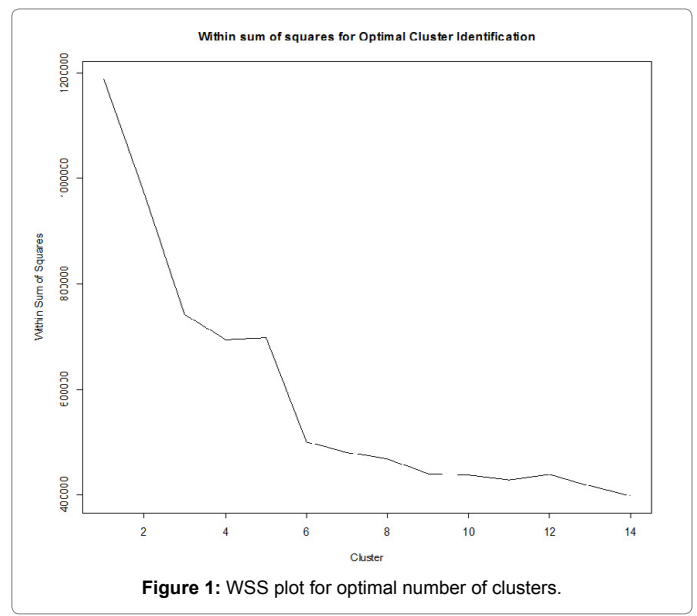

**Figure 1:** WSS plot for optimal number of clusters.

### Spectrum analysis

In vibration data when measurements of both amplitude and frequency are available, diagnostic methods can be used to determine both the magnitude of a problem and its probable cause [1]. In this research, five spectrum plots were analyzed in the time series to diagnose the state of the machine and correlate with the clusters. A frequency plot was generated for data on 12-Aug as shown in Figures 4 and 5. Based on the known maintenance history of the machine, the state was considered a normal state and was used as a baseline for the rest of the plots. The dominant cluster during this period was cluster 1.

In frequency plot generated for 27-Aug, we can observe nonsynchronous peaks through the mid band. The amplitude of the operating frequency has also significantly reduced. These characteristics can be attributed to mechanical looseness in the machine. Upon maintenance, it was discovered that the shaft had displaced by 10 mm creating mechanical looseness. The dominant cluster during this period was cluster six. In frequency plot from 10-Sep, we can observe the machine in normal operating condition comparing to the baseline as shown in Figure 6. This was the period after the maintenance. The dominant cluster during this period was cluster 2. In a frequency plot from 22-Sep, we can observe the increase in magnitude from its operating frequency as shown in Figure 7. This characteristic is an indication for mechanical imbalance. During this period the dominant cluster was cluster 4. After maintenance, we can observe the machine operating in normal condition as shown in Figure 8. The dominant cluster during this period was cluster 3.

From the above spectrum analysis, the modes of each of the clusters were diagnosed. Based on the information, some of the inferences can be drawn using the cluster plot generate using GMM. The conclusions are as follows:

- GMM model was capable of diagnosing the machine repair states. This was a clear indication of the robustness of identifying the change in process or environment.

- Imbalance state of the fan was observed since the beginning of the data collection as seen in Figure 3. Although an assumption was made during spectrum analysis while creating the baseline, clustering technique was capable of identifying the imbalance state.

- Clustering technique was capable of detecting machine powered off state as well.

From the above results it was concluded that by using clustering and spectrum analysis, it was possible to overcome some of the many challenges of supervised classification methods. Some of the advantages of the above technique were as follows:

- There was no requirement of training the model with all the states of the machine.

- The above procedure could be implemented in a shorter period. Hence, the benefit of CBM could be realized faster.

- There was no need to retrain the model when a new state of the machine is identified.

### Factor Analysis for Clustered Data

In maintenance, upon detecting and diagnosing the faults, identifying the important features that affect that are specific to a particular cluster was important. The factors contributing to a specific state of the machine was used in studying the root-cause of the problem and potentially eliminating the problem. It was also used in validating the cluster results. In this paper, a supervised learning technique called random forest was discussed. This model was used to identify the

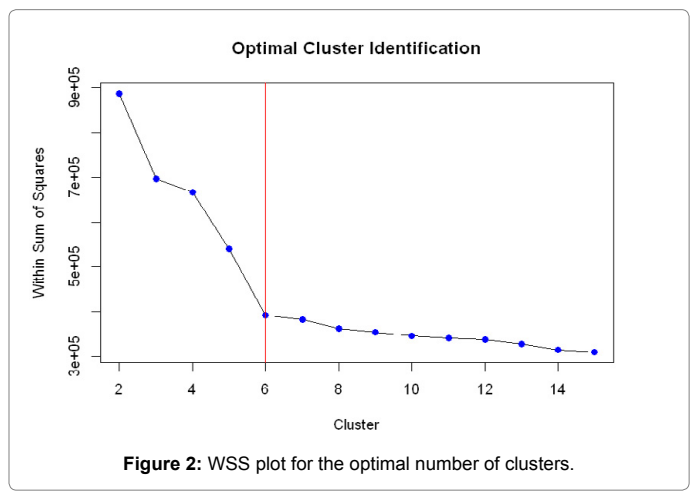

**Figure 2:** WSS plot for the optimal number of clusters.

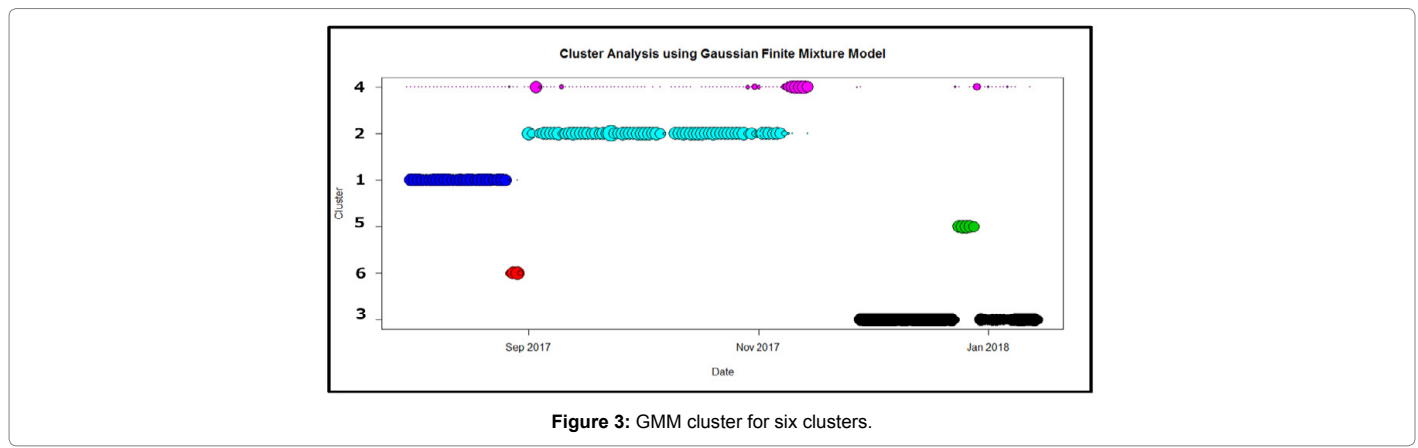

**Figure 3:** GMM cluster for six clusters.

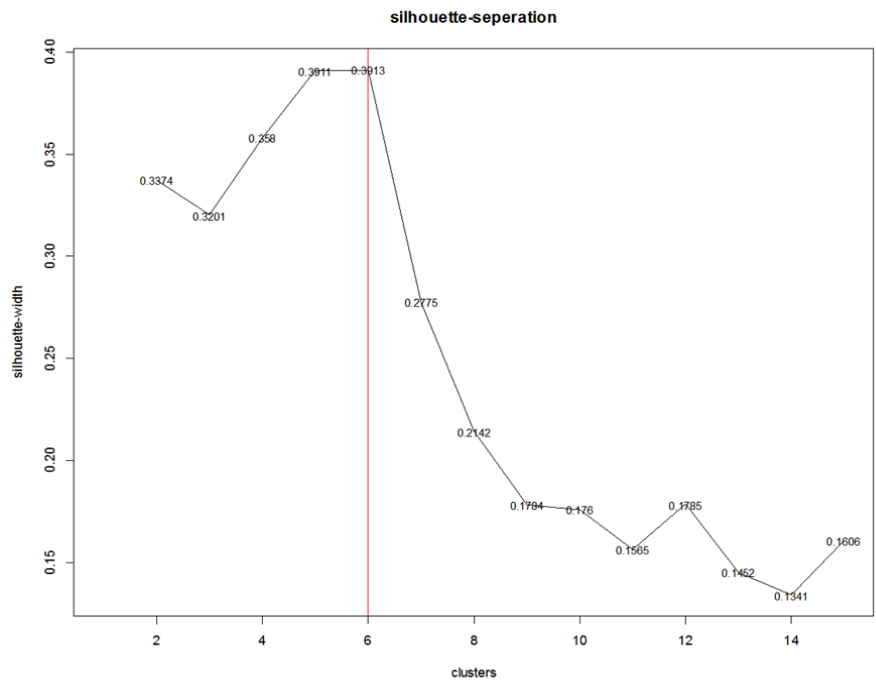

**Figure 4:** Silhouette width for identifying the optimal number of clusters.

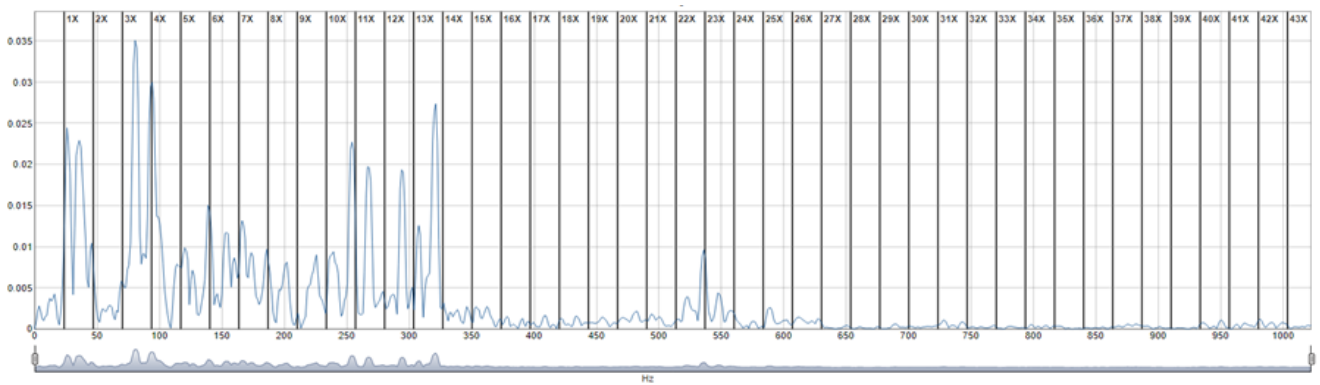

**Figure 5:** Shaft displacement issue.

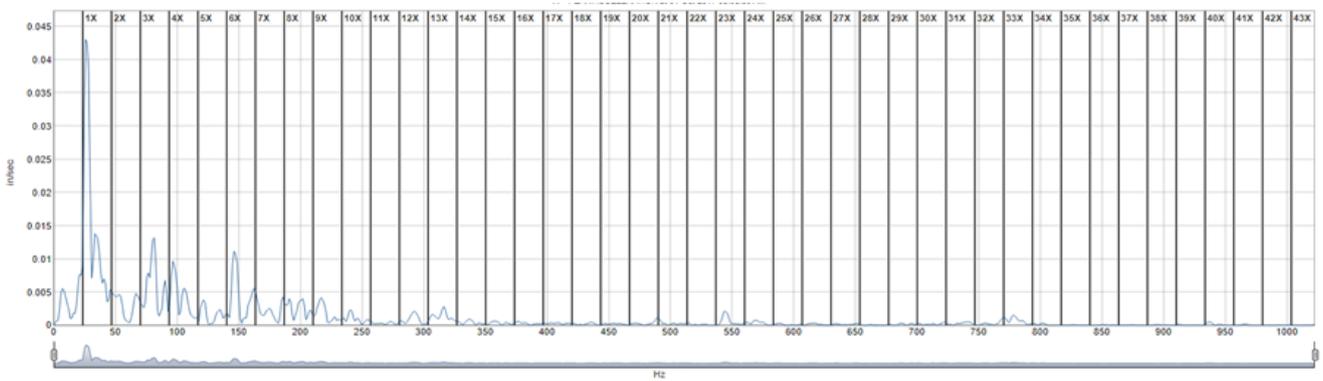

**Figure 6:** Machine normal condition or repair state.

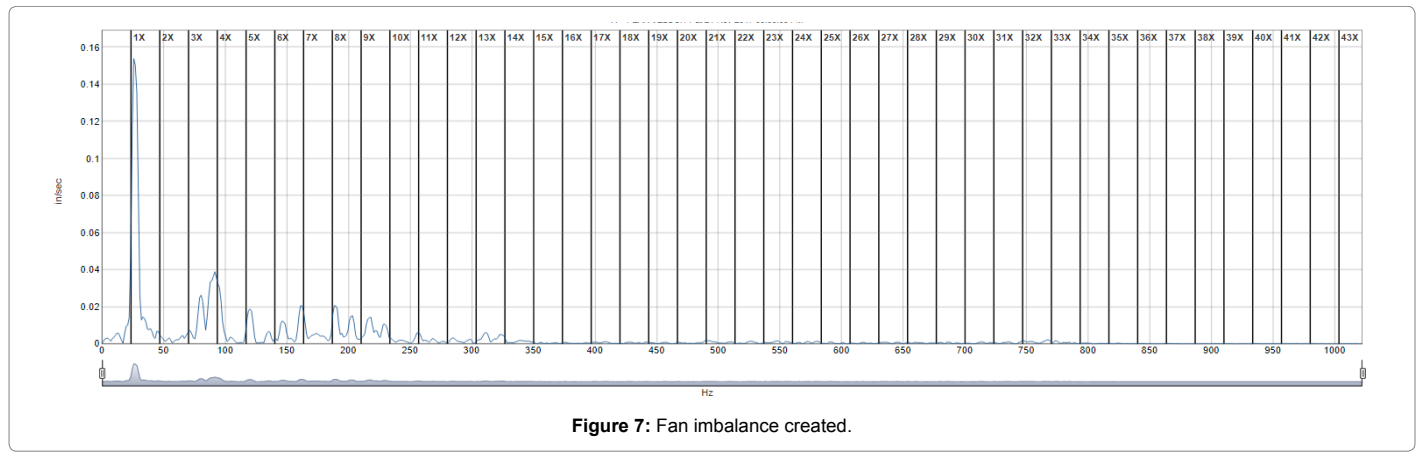

**Figure 7:** Fan imbalance created.

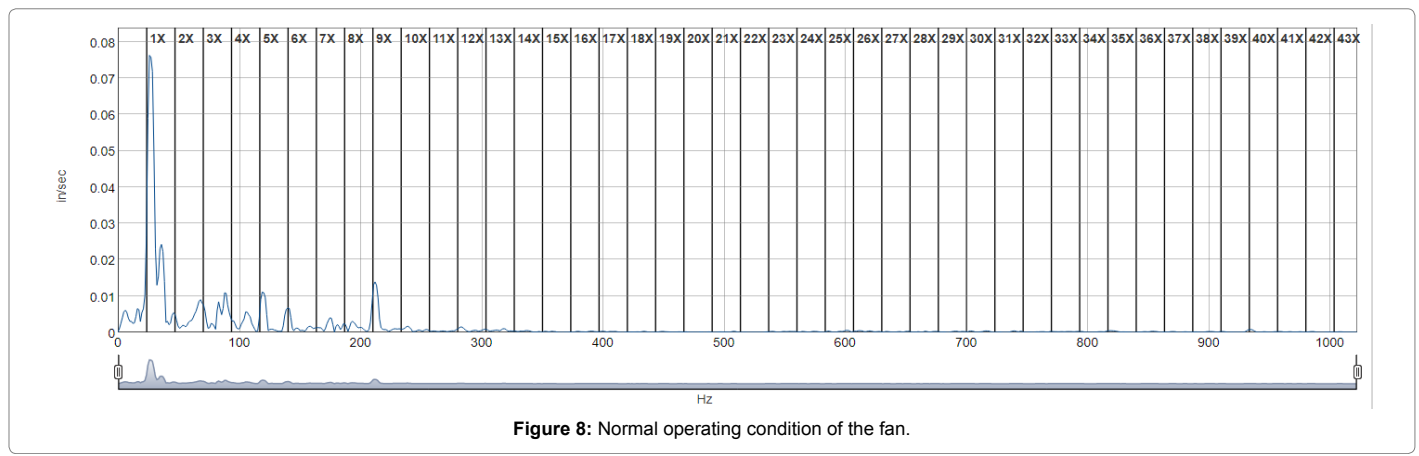

**Figure 8:** Normal operating condition of the fan.

important features that are specific to a specific fault of the machine (or cluster).

Random forest is an ensemble learning technique that is used both in regression and classification problems. In a regular decision tree, a single decision tree is built. However, in a random forest, many decision trees are built. The number of trees is usually user defined. In an ensemble process, a vote from each decision tree is used in deciding the final class. In this technique, a sample of data with replacement is used for building the decision tree along with the subset of variables. This sampling and subsetting are performed at random. Hence, this technique is called a random forest. The algorithm for the random forest is given as follows [38]:

1. Draw $n_{tree}$ bootstrap samples from the original data.

2. For each of the bootstrap samples, grow an unpruned classification or regression tree, with the following modification: at each node, rather than choosing the best split among all predictors, randomly sample $m_{txy}$ of the predictors and choose the best split from among those variables. (Bagging can be thought of as the special case of random forests obtained when $m_{txy}$=p, the number of predictors.)

3. Predict new data by aggregating the predictions of the $n_{tree}$ trees (i.e., majority votes for classification, the average for regression).

In Boosting, successive trees give extra weight to points incorrectly predicted by earlier predictors. Finally, a weighted vote is taken for prediction.

In bagging successive trees do not depend on earlier trees. Each is independently constructed using a bootstrap sample of the data. Finally, a majority vote is taken for prediction.

An estimate of the error rate can be obtained, based on the training data, by the following [38]:

1. At each bootstrap iteration, predict the data not in the bootstrap sample the tree grown with the bootstrap sample.

2. Aggregate the OOB (Out of Bag) predictions. Calculate the error rate, and call it the OOB estimate of error rate.

Variable importance in the random forest is defined based on the interaction with other variables. Random forest estimates the significance of variable based on how much the prediction error increases when data for a particular variable is permuted while the other variables are left unchanged. The calculated for variable importance are carried out each tree at a time as the random forest is constructed. Today, a random forest is used in various applications such as banking [39], retail [40], the stock market [41], medicine, gene selection [42] and image analysis [43].

Some of the main advantages of this technique are as follows:

1. The same algorithm could be used for both classification and regression problems

2. There is no issue of overfitting when this algorithm is used either for classification or regression.

3. The random forest can also be used for identifying important variables in the data while building the models.

4. It can handle large datasets efficiently without variable deletion

In this research, the clusters are used as the response variable, and the feature data is used as the predictor variables. A total of 500 trees are generated using random forest technique as shown in Figure 9. The accuracy of different models was considered to choose the best model. The optimal model was chosen with $m_{try}$=19. The summary of Resampling results across tuning parameters is as shown in Table 2.

After identifying the best model, the important variables for every cluster group was identified. The results are as shown in Figure 10.

In Figure 10, the importance's of all the features for all six clusters are shown. In the following results, it was identified that all the features had some amount of significance for cluster 1, 2, 3, 5 and 6. For cluster 4, SdYAxisF feature had no significance. We can also observe that in cluster 2, 3, 4, 5 and 6 had ambient temperature as the most significant variable.

| mtry | Accuracy | Kappa |
|------|----------|-------|
| 2 | 0.8709 | 0.8451 |
| 19 | 0.8853 | 0.8623 |
| 37 | 0.8806 | 0.8567 |

**Table 2:** Resampling results across tuning parameters.

Cluster 4 is the imbalance condition of the fan-motor assembly and temperature was the most significant factor. This explains that when an abnormal condition was observed in the fan, the temperature was a highly contributing factor in all the cases as when there is an imbalance; there was a high degree of stress on bearing increasing the temperature. Usually, motor temperature is a prime indicator of how well a motor is operating [44]. A hot motor greatly reduces the life of the unit. A 10°C (20°F) increase from the design motor temperature can reduce the life of the motor's insulation in half [45].

Cluster 1 was machine operating in a healthy condition. Here, it was observed that MeanXAxisT was the most critical feature. Usually, vibrations on axial indicate that there are vibrations due to shaft or coupling misalignment [46]. It was also important to note that during maintenance, it was identified that there was shaft displacement causing high vibration and hence confirming the vital feature. As vibration increased, the importance of MeanXAxisT remained unchanged, but the temperature significantly increased. This can be observed in cluster 6

Cluster 2 was formed after replacing the fan. Here, the most important features are temperature and SDYAxisT. Imbalance in a radial direction is mainly due to thermal problems [47]. Here we can observe that temperature was also a significant factor. This information was used to diagnose and detect thermal issues within the fan-motor assembly.

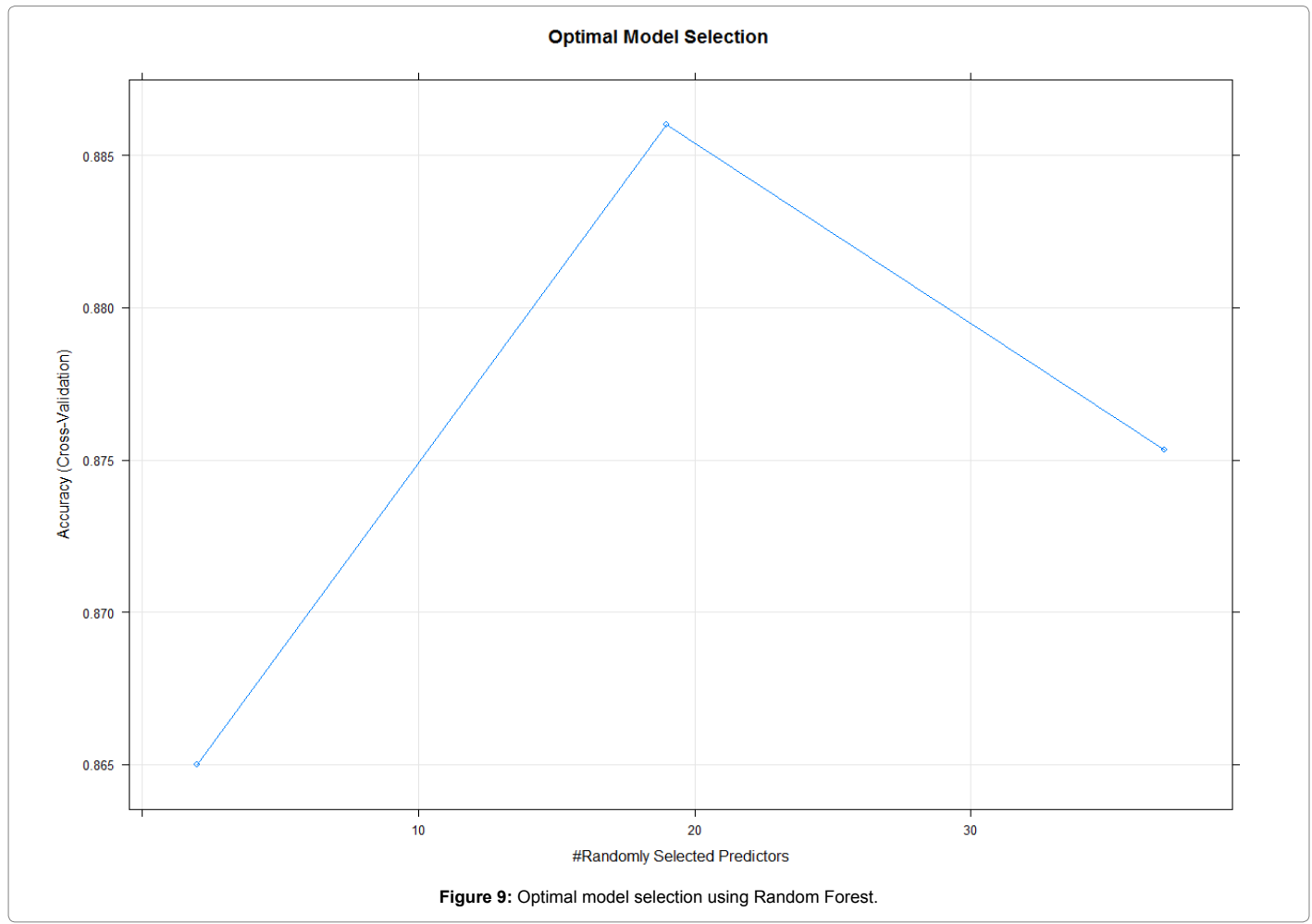

**Figure 9:** Optimal model selection using Random Forest.

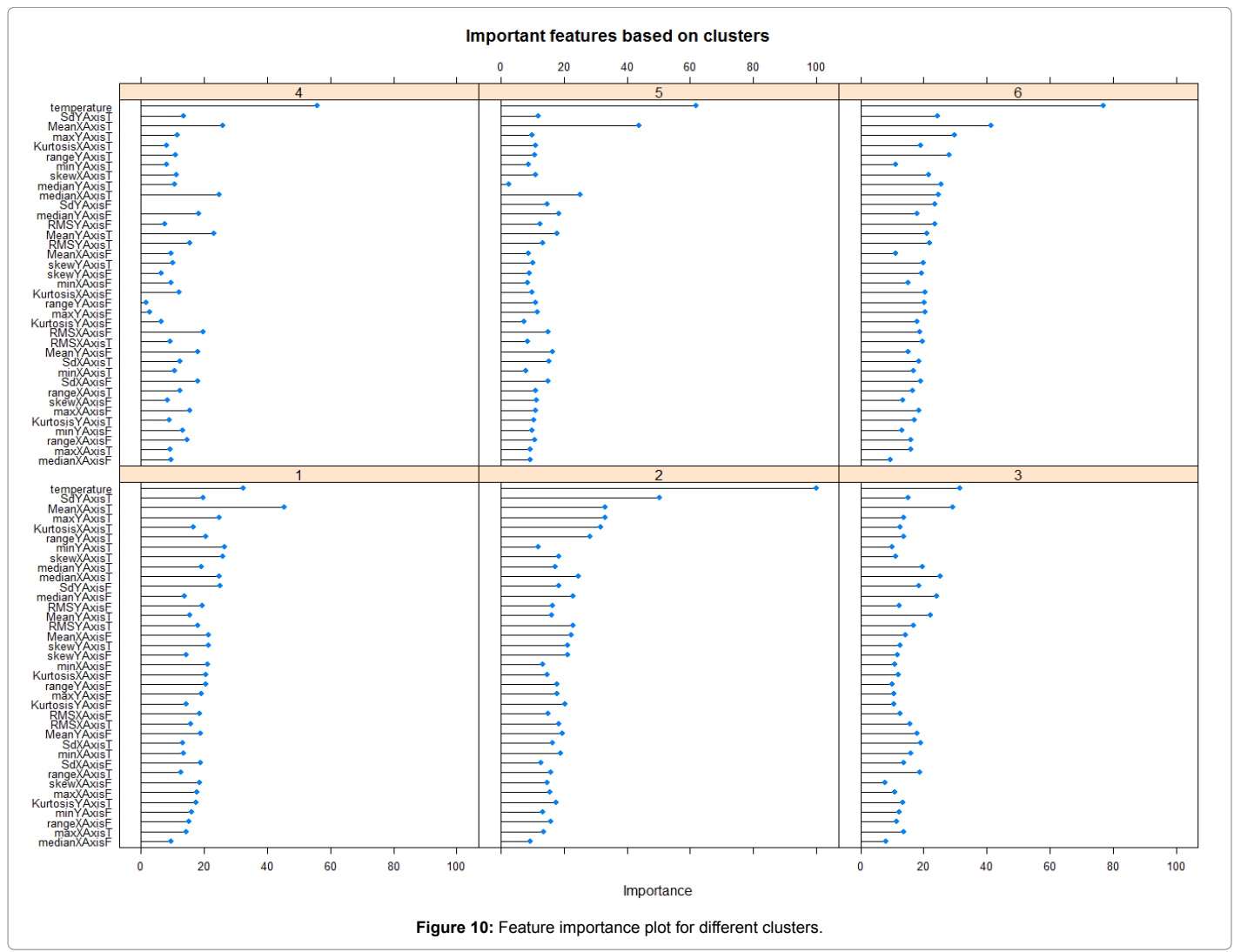

**Figure 10:** Feature importance plot for different clusters.

Finally, cluster 5 was a machines power off condition. Here, it was observed that temperature and MeanXAxisT were the most significant factors. It was also interesting to note that the importance of all the features has significantly reduced compared to other clusters. Since the fan works under a high-temperature environment, when the machine was turned off; temperature should indeed be an important factor contributing to the machine state.

From the above analysis, we can also observe that all the all the clusters have different levels of significance for different clusters. This observation provides a strong conclusion that the clusters are unique with different characteristics [48,49].

## Conclusion

In rotating machinery, vibration analysis is one of the most sought techniques for condition-based monitoring. In a highly dynamic environment such as manufacturing, unsupervised machine learning techniques such as clustering are used to group the data into clusters. These individual clusters represent a state of the machine. The mode of each state such as imbalance or bearing issues can be diagnosed using frequency spectrum analysis. The important factors that are specific to a cluster can be identified by using random forest's variable importance technique. The significant factors were used to study the causality of a particular failure mode. With different significant features for each mode, we provided substantial reasoning that the identified clusters are significantly different and their behavior was caused due to a set of unique features.

In this research, with the proposed methodology, we were able to build a machine state detection model for a machine working in changing environments using Gaussian mixture model and Expectation Maximization based clustering to eliminate the need for retraining the model, diagnosing machine faults using spectrum analysis and finally identifying important factors that contribute to each state of the machine using random forest model with its variable importance technique. The proposed research methodology was used to deduce objective reasoning for identifying factors contributing to the degradation of the machine.

In this research, there were two main objectives, and the conclusions are as follows:

***Objective 1:** to develop a technique that is capable of diagnosing different states of the machine without the need for labels*

From the results, there was sufficient rationale to conclude that using the proposed technique in this research, it was indeed possible to identify all the states of the machines without the need for labeled data.

**Objective 2:** *To identify the most critical factors that affect the causality of each state of the machine.*

Using the variable importance technique for random forest model, it was possible to identify all the critical factors that affect the causality of each state of the machine. These factors also provided strong evidence to indicate the cause and effect relationship between factors and failure modes.

## Future Work

The future scopes of this work are extended in two directions and are as follows:

### Prescriptive analytics

One of the key challenges that we come across today in maintenance is what actions must be taken to maximize the productivity of maintenance? This research is aimed to extend in this direction to use operations research technique, decision analysis, and spare part inventory control to maximize the efficiency of maintenance.

### Preemptive analysis

The ultimate goal of maintenance is to achieve breakdown free machine. This can be done by extending current work to perform a preemptive analysis where factor analysis results are used to redesign the components in the machines to achieve breakdown free machine.

### Author Biography

### Dr. Nagdev Amruthnath

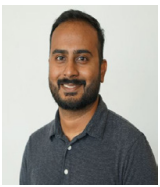

Nagdev Amruthnath is currently a Ph.D. Candidate in Industrial Engineering Department, Western Michigan University (WMU). He earned his master's degree in Industrial engineering from WMU and a Bachelor's degree in Information Science and Engineering from Visvesvaraya Institute of Technology, Karnataka, India.

He has four years of experience working in manufacturing industry specializing in the implementation of lean manufacturing, JIT technologies and production system, three years of experience in implementing data analytics, machine learning and AI technologies in manufacturing and undergraduate and graduate teaching experience. The author has journal and conference proceeding publications in production flow analysis, ergonomics, machine learning, and wireless sensor networks. Currently, his research focus is on developing machine learning and AI technologies for manufacturing application.

Nagdev continues to serve as a reviewer of scholarly journals which includes Journal of Electrical Engineering, IEEE Transactions on Reliability, and Machine Learning and Applications: An International Journal. He has also open sourced all his projects including two R-packages on GitHub for other researchers to extend his work.

Mr. Nagdev Amruthnath holds active membership in the society of manufacturing engineers. In the past, he was an active member of Indian Society for Technical Education and International Society of Occupational Ergonomics and Safety. He is also a certified in A+ &N+, and level 1 vibration analysis.

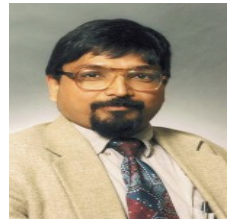

Dr. Tarun Gupta is Professor in the Department of Industrial & Entrepreneurial Engineering at Western Michigan University for 29 years. He is a Ph.D. in 1988 from the University of Wisconsin in industrial & systems engineering with a minor in computer science; a 1979 B.Tech in Mechanical Engineering from IITBHU Varanasi, and a 1981 graduate of NITIE, Mumbai India with a master's Industrial & Systems Engineering. His prime areas of research are manufacturing automation and robotics. He has also continued to research nano-sciences area, machine learning and data analytics applications in advanced manufacturing systems. He has published over 100 papers in his areas of interest. He has also been the recipient of numerous consulting assignments from area industry for specific manufacturing systems challenges.

Dr. Gupta has also served as associate editor of International Journal of Robotics and Automation and continues to serve as a reviewer of scholarly journals which includes International Journal of Production Research, International Journal of Nano photonics, International Journal of Computer & Industrial Engineering, IJCI, & Journal of Operations Management.

Dr. Gupta is a member of IEEE, lifetime member of SME and lifetime member of Society of Photonics & Instrumentation Engineers (SPIE), and a past member of IIE, ORSA, ASEE.